\documentclass{Interspeech2024}

\interspeechcameraready 

\usepackage{color,amsmath,graphicx,multirow}
\usepackage{array}
\usepackage{multirow}
\usepackage{multicol}
\usepackage{makecell}
\newcolumntype{C}[1]{>{\centering\arraybackslash}m{#1}}
\usepackage{booktabs}

\title{H\lowercase{YP}R: A comprehensive study for ASR hypothesis revising \\with a reference corpus}

\name[affiliation={1}]{Yi-Wei}{Wang}
\name[affiliation={2}]{Ke-Han}{Lu}
\name[affiliation={1}]{Kuan-Yu}{Chen}

\address{
  $^1$National Taiwan University of Science and Technology, Taiwan\\
  $^2$National Taiwan University, Taiwan}
\email{yiwei@nlp.csie.ntust.edu.tw, d12942024@ntu.edu.tw, kychen@mail.ntust.edu.tw}


\begin{document}
\maketitle

\begin{abstract}
With the development of deep learning, automatic speech recognition (ASR) has made significant progress. To further enhance the performance of ASR, revising recognition results is one of the lightweight but efficient manners. Various methods can be roughly classified into $N$-best reranking modeling and error correction modeling. The former aims to select the hypothesis with the lowest error rate from a set of candidates generated by ASR for a given input speech. The latter focuses on detecting recognition errors in a given hypothesis and correcting these errors to obtain an enhanced result. However, we observe that these studies are hardly comparable to each other, as they are usually evaluated on different corpora, paired with different ASR models, and even use different datasets to train the models. Accordingly, we first concentrate on providing an ASR hypothesis revising (HypR) \footnote{\url{https://github.com/Alfred0622/HypR}} dataset in this study. HypR contains several commonly used corpora (AISHELL-1, TED-LIUM 2, and LibriSpeech) and provides 50 recognition hypotheses for each speech utterance. The checkpoint models of ASR are also published. In addition, we implement and compare several classic and representative methods, showing the recent research progress in revising speech recognition results. We hope that the publicly available HypR dataset can become a reference benchmark for subsequent research and promote this field of research to an advanced level.
\end{abstract}

\section{Introduction}
Automatic speech recognition (ASR) systems have significantly evolved due to the growth of deep learning research in recent years. The conventional pipeline system, which mainly consists of an acoustic model, a language model, a lexicon, and a search algorithm, has gradually been phased out {\cite{jelinek1998statistical,yu2016automatic}}. End-to-end-based methods, which can directly convert input speech into text, have created a new paradigm because of their simple and elegant design \cite{LAS,Att-CTC}. To further boost the recognition performance, there are two main research directions. On the one hand, advanced model architectures and training objectives have been introduced \cite{NAR-BERT-ASR,CAKT}. On the other hand, revising the ASR output to reduce the recognition errors is a lightweight but efficient strategy \cite{MLMScore,RescoreBERT,fastcorrect}.

% \subsection{General guidelines}

The ASR hypothesis revising methods can be roughly classified into two categories: $N$-best reranking methods and error correction models \cite{Cream, chen2024hyporadise}. Since the highest-scoring recognition result may not necessarily have the fewest errors, $N$-best reranking methods thus aim to select the best hypothesis among a set of candidates for a given input speech. That is to say, this school of methods is designed to distinguish and select the most accurate hypothesis. Various methods can be further divided into token-level {\cite{MLMScore,Transcormer,ASR_Res_Tran}}, sentence-level \cite{RescoreBERT, BerlinBert}, and comparison-based strategies \cite{nie22_interspeech, DNN_Based, EC, Bert_sem}. Alternatively, an error correction model usually consists of an error detector and an error corrector. More formally, the former aims to detect recognition errors, while the latter concentrates on correcting these errors \cite{fastcorrect, EffectiveCorrect}. It is worthwhile to note that, along with the research trend, recently, more and more error correction models have attempted to blend these two components into an end-to-end approach \cite{guo2019spelling, N_Best_Correction, BART_MANDARIN,UCorrect,dutta2022error}.

% \red{All (co-)authors must be responsible and accountable for the work and content of the paper, and they must consent to its submission. Generative AI tools cannot be a co-author of the paper. They can be used for editing and polishing manuscripts, but should not be used for producing a significant part of the manuscript.}

\section{ASR Hypothesis Revising Methods}
\label{sec:RelatedWork}
To further enhance the recognition results, a number of ASR hypothesis revising methods are presented. $N$-best reranking modeling and error correction modeling are two major streams.

\subsection{$N$-best Reranking Methods}
The principal idea of $N$-best reranking methods is to determine an accurate score for each hypothesis, which should be highly negatively correlated with the error rate. That is, the higher the score, the lower the error rate. Therefore, according to the scores, the hypothesis with the highest score can be selected as the output from a set of recognition candidates generated by ASR for a given input speech. Depending on the granularity or manner in which scores are computed for each hypothesis, various approaches can be categorized into token-level, sentence-level, and comparison-based approaches.

{\bf{Token-level}} methods usually employ advanced language models to calculate the score by considering the probability of each token occurring in the hypothesis. Causal language modeling (CLM) and masked language modeling (MLM) are two popular usage schemes \cite{MLMScore}. For a given hypothesis $X=\{x_1,x_2,…,x_L\}$, causal language modeling decomposes $X$ into a sequence of conditional probabilities using the chain rule and then multiplies these probabilities to obtain the final score:
\begin{equation}
    \setlength\abovedisplayskip{3pt}%shrink space
    \setlength\belowdisplayskip{3pt}
    Score_{CLM}(X)=\prod_{i=1}^{L}{P(x_i | x_{< i})},
\end{equation}
where $x_{<i}$ denotes the partial sequence before $x_i$. Even though the calculation is mathematically sound, the probability of each token only depends on what has occurred before. However, there is no doubt that all the tokens in each hypothesis are known, especially in the context of the ASR hypothesis revising task. It is certainly possible to use bidirectional information to calculate the score for each token in the hypothesis. Therefore, in contrast to CLM, masked language modeling determines the score as follows:
\begin{equation}
    \setlength\abovedisplayskip{3pt}%shrink space
    \setlength\belowdisplayskip{3pt}
    Score_{MLM}(X) = \prod_{i=1}^{L}{P(x_i | x_{\backslash i})},
\end{equation}
where $ x_{\backslash i}$ represents the entire hypothesis except $x_i$. Compared with CLM, MLM considers more contextual information to calculate the score of each token, so it is usually expected to obtain a higher accuracy than CLM. 
 %Furthermore, MLM seems more suitable to be combined with recent Transformer-based pre-trained language models because its calculation method is consistent with the training objectives of these pre-trained models \cite{BERT}.

Typically, {\bf{sentence-level}} methods first represent each candidate using a vector representation and then convert it into a score to identify the most promising hypothesis \cite{RescoreBERT, BerlinBert}. During training, the ground truth for each hypothesis can be the error rate, the normalized ranking in a set of candidates, or the MLM (or CLM) score. By doing this, on the one hand, the computational cost is vastly reduced compared to the token-level methods during inference. This is because CLM and MLM methods have to perform calculations for each token in a hypothesis. In contrast, sentence-level methods first encapsulate a hypothesis into a vector, and then only one step is required for the hypothesis to obtain its own score. On the other hand, several off-the-shelf and powerful sentence embedding methods can be utilized to distill and encapsulate lexical, semantic, and even latent information from the hypothesis into a vector representation. By combining the advanced vector representation with task-oriented ground truth (e.g., error rate or normalized ranking) for training, a more reliable score can be obtained compared to the scores provided by token-level methods, which only consider token regularities. 

Unlike token-level and sentence-level methods that focus on calculating a score for each candidate, {\bf{comparison-based}} methods frame the $N$-best reranking task as a comparison problem. Specifically, the comparison-based models are trained to indicate the best among a set of candidates, so the absolute score of each candidate is not important \cite{nie22_interspeech}. Limited by the token numbers of model input, pairwise comparison models are most common \cite{DNN_Based,EC,Bert_sem}. Accordingly, for a set of candidates, each hypothesis should be compared with all other candidates one by one, and the hypothesis with the highest number of wins is selected as the output. Conceptually, this school of methods simplifies the reranking task from generating an accurate score for each hypothesis to determining the more correct hypothesis from a set of candidates.

\subsection{Error Correction Models}
Apart from the $N$-best reranking methods, another straightforward idea to enhance the ASR performance is to detect recognition errors and then correct these misrecognized tokens. These strategies are collectively called the error correction models, and they are usually implemented by cascading an error detection module and an error correction module in a pipeline manner \cite{fastcorrect, EffectiveCorrect, fastcorrect2, leng2023softcorrect}. As deep learning advances, upcoming research is gradually shifting toward implementation in an end-to-end and sequence-to-sequence manner. Therefore, error correction modeling can be completed in just one step, reducing the overall model complexity and avoiding possible error propagation between components. In this context, many studies attempt to leverage powerful pre-trained language models as a cornerstone to boost performance further \cite{BART_MANDARIN, dutta2022error}, while some struggle to leverage more information extracted from ASR, such as by referring to the $N$-best hypotheses \cite{N_Best_Correction}.

Several interesting observations can be made and discussed by comparing the modeling philosophies of the two streams. First, error correction modeling gives a new corrected result as the ASR output, which may not be included in the original candidate set generated by ASR. Hence, error correction modeling seems to have the potential to exceed the performance of $N$-best reranking modeling because the latter has limitations imposed by the set of the $N$-best candidates. Second, due to its end-to-end and sequence-to-sequence manner, error correction modeling may face challenges that make training difficult, while the training process of $N$-best reranking modeling is usually simpler and straightforward. Finally, without the speech utterance as a reference, error correction modeling can only detect and correct misrecognized tokens by taking into account local and global contextual textual clues. As a result, unexpected errors may occur after the error correction models are processed. In contrast, $N$-best reranking modeling appears to be a conservative but more stable alternative.

\subsection{Other Representatives and Recent Progresses}
Both the mainstream methods (i.e., $N$-best reranking modeling and error correction modeling) have their own advantages and disadvantages, and almost all of the $N$-best reranking and error correction models consider acoustic information implicitly, indirectly, or even omitted. Therefore, recent research has attempted to combine acoustic and lexical information to revise the recognition results in a cascade or an end-to-end manner \cite{Cream,cai-etal-2023-masked}. Although these approaches further improve performance, they are limited in requiring additional speech input. Moreover, there are also studies that use a word lattice as input to increase the search space and generate better recognition results \cite{10038087}. Nevertheless, encoding the topology of a word lattice as input is challenging, and these methods usually require large amounts of supervised data to train the model to leverage the lattice \cite{Neural_Lattice, LatticeBART}.

%mitigate these deficiencies and come up with a more comprehensive model by combining components of error correction modeling and $N$-best reranking modeling. An important step is to employ an ASR to determine the matching degree between each hypothesis and the given speech to avoid ill-corrections .  
Due to the rapid growth of generative large language models (LLMs), a popular paradigm in natural language processing research is zero-shot or one-shot strategies using these models, as they achieve remarkable results on various tasks. Following the research trend, their ability to handle the task of ASR hypothesis revising is of interest \cite{chen2024hyporadise, LLMforErrorCorrection, yang2023generative}. Additionally, combining an LLM with a pre-trained large-scale speech model through a cross-modal fusion technology for ASR hypothesis revising is also a novel research direction \cite{radhakrishnan-etal-2023-whispering}.
%Subsequently, in this study, we take a step forward to evaluate the giant LLMs on the ASR hypothesis revising task and make fair comparisons with other baselines under the benchmark corpus.

\section{H\lowercase{YP}R: The ASR Hypothesis Revising Benchmark}
\label{sec:dataset}

In order to promote research development and examine the progress of research, this study proposes a speech recognition hypothesis revising (HypR) benchmark. HypR includes commonly used datasets, including AISHELL-1 \cite{AISHELL1}, TED-LIUM 2 \cite{TEDLIUM2}, and LibriSpeech \cite{librispeech}. The Transformer-based CTC-attention ASR model, built using the ESPnet toolkit, is trained for each dataset \cite{ESPNET}. In more detail, the model architecture consists of an encoder and a decoder, constructed from 12 and 6 Transformers, respectively. CTC and cross-entropy losses are applied to train the model in a multi-task learning manner. No data augmentation methods are applied to enhance the performance of ASR. During decoding, the beam search algorithm is used to generate $50$ hypotheses for each speech utterance, with or without shallow fusion with an extra language model. The language models of AISHELL-1 and TED-LIUM 2 are LSTM-based models trained from scratch, while LibriSpeech applies a Transformer-based model downloaded from the ESPnet toolkit. The detailed configurations of each dataset are summarized in Table 1. In addition to the $N$-best hypotheses, checkpoints for the ASR models are also released in order to easily reproduce the contents of the benchmark, sustain development, and allow their possible use in the future.

\section{Experiments}
\label{sec:exp}
\subsection{Experimental Setup}
In the experiments, we compare the performance of several classic and representative $N$-best reranking methods and error correction models on the proposed HypR benchmark. It should be noted that without additional instructions, due to the limitations of computing resources and for a fair comparison, we used the top 10 hypotheses for each speech utterance to train, tune, and test all models in the experiments. All the hyper-parameters are selected by the development set. The model configurations, hyper-parameters, and training settings are described in detail in the repository of the benchmark. Our implementations are based on the HuggingFace library and the pre-trained language models used in this study are cloned from it \cite{HuggingFace}.

\subsubsection{$N$-best Reranking Methods}
The $N$-best reranking methods can be divided into token-level, sentence-level, and comparison-based methods. For the token-level methods, CLM and MLM are considered in our experiment. We adopt $\text{GPT2}_\text{base}$ \cite{gpt2} and $\text{BERT}_\text{base}$ \cite{BERT} to implement the CLM and MLM methods, repectively. 

RescoreBERT \cite{RescoreBERT} and PBERT \cite{BerlinBert} are selected to stand for sentence-level modeling in our experiments. The fundamental difference between them is the training objective. The former aims to predict the MLM score for each given hypothesis, while the latter targets distinguishing the hypothesis with the lowest word error rate from the others. Two auxiliary discriminative training losses for RescoreBERT, namely the minimum word error rate (MWER) and matching word error distribution (MWED), are also implemented and compared in this study \cite{RescoreBERT}.

$\text{BERT}_\text{sem}$ and $\text{BERT}_\text{alsem}$ are two comparison-based methods evaluated in this study \cite{Bert_sem}. Both are implemented using $\text{BERT}_\text{base}$ as the backbone model, and the settings follow previous research. $\text{BERT}_{\text{sem}}$ concatenates a hypothesis with one of the other hypotheses and then feeds the result into the BERT model, which is fine-tuned to reveal the degree of confidence that the former candidate may have a lower error rate than the latter. Based on $\text{BERT}_{\text{sem}}$, $\text{BERT}_{\text{alsem}}$ is a variant that not only stacks Bi-LSTM layers on top of BERT to obtain better representations but also considers the scores calculated by ASR to try to provide a more indicative score. Hence, the major difference between them is the statistics used in the model. $\text{BERT}_\text{sem}$ only considers textual information, while $\text{BERT}_\text{alsem}$ takes acoustic, linguistic, and textual information into account.

Inspired by $\text{BERT}_\text{alsem}$, which introduces LSTM layers on top of BERT to obtain better representations, we stack a Bi-LSTM layer on top of BERT for the PBERT model. Therefore, the contextualized token embeddings generated by the last layer of BERT would also go through the Bi-LSTM layer. After that, average and max pooling are performed to collect token-level clues. The two pooling vectors, the hypothesis embedding generated by BERT (i.e., the embedding of the special [CLS] token), and the ASR decoding score are concatenated to represent the target hypothesis. We denote the simple extension as PBERT+LSTM hereafter. The training objectives of PBERT and PBERT+LSTM are exactly the same. 

\subsubsection{Error Correction Models}
BART is a widely used end-to-end and sequence-to-sequence model, so we choose it as a representative of error correction modeling \cite{BART_MANDARIN,dutta2022error}. During training, BART aims to convert each input hypothesis into its corresponding ground truth, while only the top hypothesis is used to generate the final output for a given speech during the inference stage. Based on BART, we further implement $N$-best Plain and $N$-best Alignment methods to access more information from $N$-best hypotheses to try to make more accurate corrections \cite{N_Best_Correction}. For both methods, we use the top four hypotheses for realizing, following the setting of the original study.

\subsubsection{Leveraging Large Language Models}
As we lack direct access to the model parameters of ChatGPT, we employ a methodology known as prompt engineering to assess its zero-shot capabilities in this study. We interact with ChatGPT through the official API\footnote{gpt-3.5-turbo-0613, \url{https://api.openai.com}} and ask it to select or generate the most probable response from a list of $10$ hypotheses. For the former, we ask ChatGPT to output a number indicating which of the hypotheses in the candidate set is selected as the output. Such a strategy limits the results to the predefined set of candidates. In contrast, for the latter, ChatGPT is asked to generate a response by considering the information provided by a set of candidates, allowing for a more open-ended result.

%\subsubsection{Evaluation settings}
%As for the hyperparameters in our experiments, we search through the hyperparameters $\alpha_{ASR}, \alpha_{CTC}, \alpha_{LM}, \alpha_{Rescore}$ in the range [0.0, 1.0] as the weight of score from ASR, CTC, language model applied during shallow fusion and rescore model respectively.  The final scores of candidate sentences are decided by the following formula:
%\begin{align*}
%    Score(X) & = \alpha_{ASR} \times Score_{ASR}(X) + \alpha_{CTC} \times Score_{CTC}(X) \\
%             & + \alpha_{LM} \times Score_{LM}(X) + \alpha_{Rescore} \times Score_{Rescore}(X) \\ 
%\end{align*}
%We find the best combination of weights for each rescore model during evaluation, and the weights are searched by using the development set of each dataset.

 \begin{table}[]{
\scriptsize
\centering
 \begin{tabular}{m{2.5cm}C{1.3cm}C{1.3cm}C{1.3cm}}
 \Xhline{1pt}
  & \multicolumn{1}{c}{AISHELL-1} & \multicolumn{1}{c}{TED-LIUM 2} & LibriSpeech \\ \hline\hline
 Training Epochs & \multicolumn{1}{c}{50} & \multicolumn{1}{c}{50} & 100 \\ 
 Batch Size & \multicolumn{1}{c}{64} & \multicolumn{1}{c}{32} & 128 \\ 
 Learning Rate & \multicolumn{1}{c}{1.0} & \multicolumn{1}{c}{2.0} & 10.0 \\ 
 Grad-Accum & \multicolumn{1}{c}{2} & \multicolumn{1}{c}{2} & 2\\
 Warmup Steps & \multicolumn{1}{c}{25,000} & \multicolumn{1}{c}{25,000} & 25,000\\
 Vocab Size & \multicolumn{1}{c}{4,233} & \multicolumn{1}{c}{502} & 5,002 \\ 
 CTC Weight (Training) & \multicolumn{1}{c}{0.3} & \multicolumn{1}{c}{0.3} & 0.3\\ 
 CTC Weight (Decoding) & \multicolumn{1}{c}{0.5} & \multicolumn{1}{c}{0.3} & 0.4 \\ 
 LM Weight (Decoding) & \multicolumn{1}{c}{0.7} & \multicolumn{1}{c}{0.5} & 0.7 \\
 %$\text{Weight}^{train}_{CTC}$ & \multicolumn{1}{c}{0.3} & \multicolumn{1}{c}{0.3} & 0.3\\ 
 %$\text{Weight}^{decode}_{CTC}$ & \multicolumn{1}{c}{0.5} & \multicolumn{1}{c}{0.3} & 0.4 \\ 
 %$\text{Weight}^{decode}_{LM}$ & \multicolumn{1}{c}{0.7} & \multicolumn{1}{c}{0.5} & 0.7 \\
 \Xhline{1pt}
 \end{tabular}
 }
 \caption{Detailed configurations of the ASR systems.}
 \label{tab:ASR_setup}
 \vspace{-3em}
 \end{table}

%----------------------------

\begin{table*}[h]
\scriptsize
\centering
\setlength{\tabcolsep}{1.3pt}
\begin{tabular}{
m{2.8cm}
C{0.7cm}C{0.7cm}C{0.7cm}C{0.7cm}
C{0.7cm}C{0.7cm}C{0.7cm}C{0.7cm}|
C{0.7cm}C{0.7cm}C{0.7cm}C{0.7cm}
C{0.7cm}C{0.7cm}C{0.7cm}C{0.7cm}
}
\Xhline{1pt}
    & \multicolumn{8}{c|}{w/o LM} & \multicolumn{8}{c}{w/ LM}\\
    \hline
    & \multicolumn{2}{c}{AISHELL-1} & \multicolumn{2}{c}{TED-LIUM2} & \multicolumn{4}{c|}{LibriSpeech} & \multicolumn{2}{c}{AISHELL-1} & \multicolumn{2}{c}{TED-LIUM2} & \multicolumn{4}{c}{LibriSpeech} \\
    & \multirow{2}{*}{Dev} & \multirow{2}{*}{Test} & \multirow{2}{*}{Dev} & \multirow{2}{*}{Test} & \multicolumn{2}{c}{Dev} & \multicolumn{2}{c|}{Test} & \multirow{2}{*}{Dev} & \multirow{2}{*}{Test} & \multirow{2}{*}{Dev} & \multirow{2}{*}{Test} & \multicolumn{2}{c}{Dev} & \multicolumn{2}{c}{Test} \\
    &&&&& Clean & Other & Clean & Other &&&&& Clean & Other & Clean & Other \\
    \hline\hline
    \textit{Baseline} &&&&&&&&&&&&&&&\\
    Top1 & 7.34 & 8.33 & 14.29&12.07&4.44&11.94&4.64&12.00&6.66&7.23&11.74&9.72&2.55&7.12&2.85&7.21\\
    Oracle &3.98&4.74&11.49&8.55&2.28&8.17&2.43&8.14&3.70&4.14&9.05&6.28&1.15&4.48&1.35&4.45\\
    Random &11.68&12.39&15.50&14.22&7.90&14.66&8.10&14.79&11.26&11.64&13.14&12.13&6.49&11.07&6.80&11.10\\
    \hline
    \textit{N-best Reranking} &&&&&&&&&&&&&&&\\
    \ \ \textit{Token-level} &&&&&&&&&&&&&&&\\
    \quad CLM &6.05&6.80&13.06&10.61&3.42&9.89&3.64&9.99&5.92 &6.55& 11.17& 8.92& 2.50& 6.86& 2.84& 7.05\\
    \quad MLM\cite{MLMScore} &5.17&6.03&12.75&10.39&3.31&9.77&3.59&9.81&4.98&5.58&10.64&8.58&2.39&6.74&2.76&6.91\\
    \ \ \textit{Sentence-level} &&&&&&&&&&&&&&& \\
    \quad RescoreBERT\cite{RescoreBERT} &5.29&6.12&13.10&10.61&3.41&9.91&3.69&9.98&4.99&5.56&11.16&8.77&2.38&6.83&2.79&6.96\\
    \quad \quad +MWER &5.29&6.11&13.03&10.60&3.40&9.89&3.68&9.93&4.99&5.59&10.90&8.67&2.39&6.80&2.79&6.94\\
    \quad \quad +MWED & 5.30&6.10&13.02&10.66&3.37&9.91&3.66&9.97&5.06&5.62&11.06&8.83&2.38&6.81&2.79&6.95\\
    \quad PBERT\cite{BerlinBert} &5.08&5.89&12.81&10.35&3.36&9.83&3.63&9.97&4.76&5.35&10.49&8.51&2.35&6.69&2.74&6.81\\
    \quad PBERT+LSTM &5.04&5.84&12.72&10.25&3.34&9.82&3.58&9.90&4.71&5.30&10.46&8.31&2.35&6.50&2.64&6.66\\
    \ \ \textit{Comparison-based} &&&&&&&&&&&&&&&\\
    \quad $\text{BERT}_\text{sem}$ \cite{Bert_sem} &5.27&6.05&13.08&10.53&3.46&10.18&3.83&10.20&5.00&5.63&10.80&8.85&2.50&6.84&2.85&6.96\\
    \quad $\text{BERT}_\text{alsem}$ \cite{Bert_sem} &5.25&6.08&12.94&10.55&3.49&10.12&3.79&10.16&5.31&5.95&10.99&9.12&2.45&6.89&2.83& 6.89\\
    \hline
    \textit{Error Correction} &&&&&&&&&&&&&&&\\
    \quad BART &6.62&7.44&13.51&11.77&4.21&11.02&4.50&11.26&6.94&7.53&12.32&10.66&2.71&7.26&3.05&7.42 \\
    \quad\quad +4-best Plain \cite{N_Best_Correction} &7.10&8.10&14.14&11.81&4.24&11.60&4.44&11.69&6.38&6.99&11.96&10.04&2.69&7.21&3.00&7.37 \\
    \quad\quad +4-best Alignment \cite{N_Best_Correction} &7.37&8.35&14.89&12.99&4.58&11.97&4.81&12.09&6.76&7.31&12.44&10.50&2.85&7.40&3.18&7.64 \\
    \hline
    \textit{Using LLM} &&&&&&&&&&&&&&&\\
    \quad Selection&6.57&7.36&13.79&11.67&4.79&11.07&4.90&11.33&8.16&8.60&12.38&11.15&5.44&9.57&5.72&9.72 \\
    \quad Generation & 6.67&7.71&16.34&13.95&5.59&12.19&6.07&12.77&6.97&7.40&14.84&14.02&5.52&9.86&6.01&10.70\\
    \Xhline{1pt}
\end{tabular}
\caption{Experimental results of various classic and representative methods on the HypR benchmark.}
\label{tab:result_all}
\vspace{-3.5em}
\end{table*}

\subsection{Performance Analysis}
Table \ref{tab:result_all} summarizes the word error rate performance for various methods on the proposed HypR benchmark. Additionally, we take the AISHELL-1 corpus as an example to report the model sizes of these methods and calculate the average computation time (in seconds) they are required to generate a final output. These statistics are listed in Table \ref{tab:param_num}. Based on these results, some interesting observations and results worth exploring can be obtained. First, focusing on baseline systems, we can conclude that using language models for decoding does improve the ASR performance. Among the different datasets, the performance gap for LibriSpeech is particularly large. A possible reason for this is that the language model used for LibriSpeech is a Transformer-based model, which is more robust than the LSTM-based models used for the other datasets. By comparing the differences between the Oracle and Top1 systems, we can see that there is still a long way to go to improve recognition results, which also illustrates the value of this line of research. When comparing Oracle and Random systems, the experiments show significant differences for all datasets, demonstrating the diversity of the $N$-best hypotheses in HypR.

Next, we turn to examining the $N$-best reranking methods. At first glance, although there is not much difference in the performance of various methods, sentence-level modeling seems to be the best choice. Looking at the results in more detail, for token-level modeling, MLM achieves better results than CLM, as expected. The reason for this might be that MLM considers bidirectional information to determine the score for each token in the hypothesis, while CLM only takes unidirectional clues into account. It is worth noting that although CLM and MLM have almost the same computational cost in theory, CLM is faster than MLM in practice. This is because CLM can be implemented with the attention masking strategy to reduce computation costs associated with iterative decoding, but MLM cannot do so because of information leakage \cite{Transcormer,shin2020fast}. For that reason, CLM is more than ten times faster than MLM in our experiments. Next, we compare RescoreBERT, PBERT, and PBERT+LSTM, which are all sentence-level methods. Since MLM scores are not precisely positively correlated with obtaining the lowest error rate, there is no doubt that PBERT and PBERT+LSTM could obtain better results than RescoreBERT. There are two more notable observations. On the one side, the efficiency of RescoreBERT combined with the MWER or MWED does not appear in HypR. On the other side, it should be emphasized that our proposed simple extension PBERT+LSTM is better than other sentence-based methods and shows superior results compared to all other methods. However, due to the addition of the Bi-LSTM layer, PBERT+LSTM not only has more model parameters but is also slower than PBERT and RescoreBERT. Finally, comparison-based methods, including $\text{BERT}_\text{sem}$ and $\text{BERT}_\text{alsem}$, have a high computational cost and cannot provide better results than token-level and sentence-level methods. Although $\text{BERT}_\text{alsem}$ considers acoustic and language model scores while $\text{BERT}_\text{sem}$ does not, the former only achieves better results on LibriSpeech, especially when paired with a language model for generating $N$-best hypotheses. 

In the third set of experiments, we consider error correction modeling. BART and two extensions are examined in the experiments. Surprisingly, both extensions do not always deliver better results than BART. BART obtains better results than the two extensions in the "w/o LM" setting, and BART+4-best Plain outperforms BART and BART+4-best Alignment if the quality of the $N$-best hypotheses is at a certain level (i.e., w/ LM). The results are somewhat reasonable because these two extensions mainly focus on leveraging more hypotheses to enhance the results, so the quality of the hypotheses will definitely affect the results.

In the last set of experiments, we are interested in the results obtained using ChatGPT. Even though ChatGPT is a generative model with a well-known advantage in performing tasks related to natural language generation, the hypothesis selection strategy produces better results than the generation strategy in most cases. We also observe that the Selection and Generation strategies perform even worse than the Top1 system in the case of decoding with a language model. The reason for this may be that the recognition candidates generated by fusing a language model are fluent, reasonable, and grammatical. Therefore, it is difficult for ChatGPT to select or generate better results. By comparing LLM-based methods with other models, the experiments show that finding better ways to leverage large language models to revise the ASR hypothesis remains urgent.

%\begin{table}[]{
% \begin{tabular}{m{2.1cm}C{3.5cm}}
% \Xhline{1pt}
%  & \multicolumn{1}{c}{Model Size}  \\ \hline\hline
% CLM & 102.07M \\ 
% MLM & 102.29M \\ 
% RescoreBERT & 102.27M\\
% $\text{BERT}_\text{sem}$  & 102.27M\\
% $\text{BERT}_\text{alsem}$  & 109.26M\\
% BART & 140.19M\\
% \quad + nBest Align & 142.55M \\ 
% PBERT & 102.27M \\
% \quad + LSTM & 140.44M \\
% \Xhline{1pt}
% \end{tabular}
% }
% \caption{The parameter size of each model}
% \label{tab:param_num}
% \end{table}

 %%%%%%%%%%%%%%%%%%%%%%%%%%%%%%%%%%%%%%%

\begin{table}[]{
\scriptsize
\centering
 \begin{tabular}{m{2.5cm}C{1.8cm}C{1.8cm}}
 \Xhline{1pt}
  & \multicolumn{1}{c}{Model Parameters} & \multicolumn{1}{c}{Latency}  \\ \hline\hline
  \textit{N-best Reranking} &&\\
   \ \ \textit{Token-level} &&\\
        \quad CLM & 102.07M & 0.014 \\
        %\quad \quad +AR Decoding & & 0.203 \\
        \quad MLM & 102.29M & 0.185 \\ 
   \ \ \textit{Sentence-level} &&\\
        \quad RescoreBERT & 102.27M & 0.013\\
        \quad PBERT & 102.27M & 0.014\\
        \quad PBERT+LSTM & 140.44M & 0.020\\
   \ \ \textit{Comparison-based} &&\\
        \quad $\text{BERT}_\text{sem}$  & 102.27M & 0.138\\
        \quad $\text{BERT}_\text{alsem}$  & 109.26M & 0.134\\
\hline
   \textit{Error Correction} &&\\
        \quad BART & 140.19M & 0.186\\
        \quad\quad + 4-best Plain & 140.19M & 0.191 \\
        \quad\quad + 4-best Alignment & 142.55M & 0.212\\ 
 \Xhline{1pt}
 \end{tabular}
 }
 \caption{The number of model parameters and decoding time (in seconds) for various models.}
 \label{tab:param_num}
 \vspace{-3em}
 \end{table}
 
%\begin{table}[]{
% \begin{tabular}{m{2.1cm}C{3.5cm}}
% \Xhline{1pt}
%  & \multicolumn{1}{c}{AISHELL-1 Test}  \\ \hline\hline
% CLM & 0.014 \\ 
% MLM & 0.183 \\ 
% RescoreBERT & 0.013\\
% $\text{BERT}_\text{sem}$  & 0.138\\
% $\text{BERT}_\text{alsem}$  & 0.134\\
% BART & 0.206\\
% \quad + nBest Align & 0.179 \\ 
% PBERT & 0.014 \\
% \quad + LSTM & 0.020 \\
% \Xhline{1pt}
% \end{tabular}%
% }
% \caption{The real time factor of each model}
% \label{tab:RTF}
% \end{table}

\section{Conclusion}
\label{sec:conclusion}
This paper presents a publicly available ASR hypothesis revising (HypR) benchmark, and a series of experiments are conducted on the corpus to perform fair analyses and comparisons. Several observations and discussions are also addressed in this study. In the future, we plan to enrich and diversify the benchmark. This will include utilizing different ASR models like RNN-T and Whisper to generate hypotheses, examining the revising performance on domain-specific corpora such as AMI and SEMEAN datasets, and more.

\section{Acknowledgement}
This work was supported by the National Science and Technology Council of Taiwan under Grants NSTC 112-2636-E-011-002 and NSTC 112-2628-E-011-008-MY3. We thank the National Center for High-performance Computing of the National Applied Research Laboratories (NARLabs) in Taiwan for providing computational and storage resources.

\bibliographystyle{IEEEtran}
\bibliography{mybib}

\end{document}